\documentclass[10pt]{article}
\usepackage[letterpaper]{geometry}
\usepackage{hicss}
\usepackage{times}
\usepackage[none]{hyphenat}
\usepackage{url}
\usepackage{latexsym}
\usepackage{minted}
\usepackage{indentfirst}
\usepackage{graphicx}
\graphicspath{{images/}}
\usepackage[
    style=apa,
  ]{biblatex}
\usepackage{xcolor}
\usepackage{amssymb}
\usepackage{algorithm}
\usepackage{algorithmic}
\usepackage{bbding}
\usepackage{pifont}
\usepackage{multirow}
\usepackage{booktabs}

\usepackage{amssymb}
\usepackage{amsthm,multicol,enumerate}
\usepackage{makecell}
\usepackage{graphicx}
\usepackage{color,xcolor}
\usepackage{subcaption}
\usepackage{enumitem}
\usepackage{soul}
\usepackage{array}
\usepackage{multirow}
\usepackage{amsmath}
\usepackage{tabularx}

\title{Quantifying True Robustness: Synonymity-Weighted Similarity for Trustworthy XAI Evaluation}

\author{Christopher Burger \\
 The University of Mississippi \\
 {\underline{ cburger@olemiss.edu}}} 
\date{}

\begin{document}
\maketitle
\begin{abstract}

Adversarial attacks challenge the reliability of Explainable AI (XAI) by altering explanations while the model's output remains unchanged. The success of these attacks on text-based XAI is often judged using standard information retrieval metrics. We argue these measures are poorly suited in the evaluation of trustworthiness, as they treat all word perturbations equally while ignoring synonymity, which can misrepresent an attack's true impact. To address this, we apply synonymity weighting, a method that amends these measures by incorporating the semantic similarity of perturbed words. This produces more accurate vulnerability assessments and provides an important tool for assessing the robustness of AI systems. Our approach prevents the overestimation of attack success, leading to a more faithful understanding of an XAI system’s true resilience against adversarial manipulation.

\end{abstract}

\subsubsection*{Keywords:}

Explainability, Robustness, Similarity, Stability, XAI
\newline

\section{Introduction}
Intractably complex models have proliferated in recent years as a result of their efficacy over their simpler relatives. And while efficacy is a necessary component of a good model, it is difficult to trust a black-box and so the question of ``\textit{How does this model actually work?}’’ follows. The lack of answers to this question has slowed the adoption of powerful models in fields where flawed models can have severe consequences, such as medicine \parencite{markus2021role,tjoa2020survey}. 

The discipline of Explainable AI (XAI) attempts to answer our question by generating an explanation of how the model generates its output. However, the XAI process that generates the explanation is itself a model, and while by design the XAI model can be inherently understood, we return to the fundamental question of any model ``\textit{How effective is it?}’’ Unfortunately, this question is nearly as difficult as our original query. As the models we attempt to explain are themselves incapable of being (comprehensively) understood, we have no ground truth against which to compare the output of the XAI method.

However, we can judge the quality of an XAI method using criteria that avoid an understanding of the original intractable model and instead focus on the consistency of the output of the XAI method when subject to perturbations that retain the meaning of the original input. The property is called stability, or robustness, and is a necessary attribute required to trust an explanation. In general, stability is the intuitive property where insignificant changes to the input lead to only small changes to the output. An XAI process that lacks stability produces substantially different explanations under similar inputs and so is of limited, if any, use. 

To assess stability, it is necessary to compare the output of the explainability method, and in doing so, some measure is required to provide the comparisons. These similarity measures are the engine that drives the adversarial search process and so suboptimal measures will result in incorrect conclusions of an XAI methods stability. The similarity measures here are subject to two primary flaws. The first is the \textit{ sensitivity}, where two explanations are fundamentally similar, but minor differences are drastically magnified, resulting in effective false positive attack success. The second is the opposite, \textit{indifference} or \textit{coarseness} where important differences between the explanations are not captured.  Although the measures and metrics in XAI have been directly discussed in previous research \parencite{hoffman2019metrics}, the focus was primarily on evaluating the quality of the explanations rather than the quality of the comparison between the explanations. 
In particular, the work \cite{burger2023explanations} established desiderata in text-based adversarial XAI using common desirable attributes in information retrieval \parencite{GenDistKumar}, and used these to motivate an algorithm that mitigates the previously mentioned flaws. However, no prior work on text-based XAI has built purpose-driven measures and instead relied on standard formulations.

Here, we elect to alter the similarity measure directly using synonymity weighting, where a perturbed word and its original are compared using an inner measure that incorporates their synonymity. Standard measures operate on strict equality between features (words in our case) in the lists being compared. This discards valuable information contained within the structure of the language itself. The removal of this information is not congruent with our focus of maintaining close meaning of the perturbed word, the similarity measure should reflect the end goal of the adversarial process by accounting for the ``closeness'' between the words. 

Table \ref{tab:comparative_perturbations_example} provides an example of how standard measures tend to be overly sensitive despite the explanations being intuitively very similar. 
We see large increases in similarity under a measure that uses synonymity weighting on perturbed features, which follows our intuition, as the words fulfill the same structure and possess much of the same meaning. Our focus on altering the similarity measures also has the advantage of reduced complexity in that the search process does not need to be taken into account in the attack algorithm. Since existing measures tend towards the sensitive side, careful choice in which words are selected for perturbation is required. Without this limitation, we can now freely replace the search process, which also gives us the ability to use many adversarial natural language processes directly which otherwise would require modifications such as those in \cite{burger2023explanations}.

Our contributions are: (1) The extension of some common measures in information retrieval to use synonymity weighting, focused on providing a more appropriate choice for adversarial methods in XAI. We leverage prior works in element-weighted similarity to allow a more accurate comparison between explanations which provides for a superior understanding of an XAI method's robustness. (2) A comprehensive comparison against adversarial examples previously determined to be successful at some threshold of similarity under the standard measures. We show that prior conclusions about XAI instability drawn using certain existing measures may be inaccurate, especially for methods with naive search procedures. Ultimately, without a reliable way to measure the impact of adversarial perturbations, we cannot establish the trustworthiness of an explanation, which is a critical prerequisite for AI Safety in high-stakes domains.

\begin{table}[tb!]
    \caption{Comparative similarities between the original and synonymity weighted versions of common measures on an adversarial example of the Symptoms to Diagnosis dataset}
    \footnotesize

    \begin{center}
     \textul{\textbf{Original Text} }  
    \end{center}
    
    I've been feeling really sick and I have a rash all over my body. I'm worried about what it could be.
    \newline
    
    \begin{center}
     \textul{\textbf{Perturbed Text}}   
    \end{center}

    I've been feeling \textbf{\textit{real}} \textbf{\textit{sickly}} and I have a rash all over my body. I'm \textbf{\textit{alarmed}} about what it could be.
\vspace{10pt}

    \begin{subtable}[t]{.48\textwidth}
    \raggedright
    \setlength\tabcolsep{11.5 pt}
        \begin{tabular}{clc|clc}
           \multicolumn{3}{c}{\textbf{\textul{Original Explanation}}} & \multicolumn{3}{c}{\textbf{\textul{Perturbed Explanation}}} 
            \\
             \multicolumn{2}{c}{\textbf{}} & \multicolumn{1}{c}{\textbf{}}&  \multicolumn{2}{c}{\textbf{}} & \multicolumn{1}{c}{\textbf{}}  \\
            \hline \\
            1 & rash &  & 1 & body &   \\
                2 & body &  & 2 & rash &  \\
                3 & worried &  & 3 & \textbf{\textit{alarmed}} &  \\
                4 & really &  & 4 & feeling &  \\
                5 & sick &  & 5 & \textbf{\textit{sickly}} &  \\
                6 & feeling &  & 6 & over &  \\
                7 & over & & 7 & \textbf{\textit{real}}  & \\
                
            \hline \\
        \end{tabular}
        \end{subtable}

\renewcommand{\tabcolsep}{2.5pt}
\centering
\begin{subtable}[t]{.48\textwidth}
\caption*{\textbf{Comparative Similarities}}
\raggedright
\begin{tabular}{cccccc}
\toprule

\multicolumn{6}{c}{\textbf{\textul{Standard}}} \vspace{4pt} \\

\textbf{Jaccard} & \textbf{Kendall} & \textbf{Spearman} & \textbf{RBO$_{0.5}$} & \textbf{RBO$_{0.7}$} & \textbf{RBO$_{0.9}$}\\
0.40 & 0 & 0.35 & 0.40 & 0.48 & 0.54

\\

\multicolumn{6}{c}{\textbf{\textul{Weighted}}} \vspace{4pt} \\
\textbf{Jaccard} & \textbf{Kendall} & \textbf{Spearman}
& \textbf{RBO$_{0.5}$} & \textbf{RBO$_{0.7}$} & \textbf{RBO$_{0.9}$}\\
0.64 & 0.23 & 0.59 & 0.88 & 0.89 & 0.85\\

\bottomrule
\end{tabular}
    \end{subtable}%

    \label{tab:comparative_perturbations_example}
\end{table}









\section{Background and Related Work}
\label{sec:motivation}
Prior work on XAI stability has focused on evaluating models using tabular or image data in various interpretation methods, which often use small perturbations of the input data to generate appreciably different explanations \parencite{alvarezmelis2018robustness,InterpretationNN,alvarezmelis2018robustness}, or generate explanations consisting of arbitrary features \parencite{slack2020fooling}. As our goal is to leverage the idea of \textit{synonymity} inherent to features of an XAI explanation, we concentrate exclusively on XAI methods that operate on text-based input.  Previous work directly involves adversarial perturbations for text-based XAI using a variety of search processes and similarity measures / distance metrics \parencite{sinha2021perturbing,ivankay2022fooling,burger2023explanations, burger2025}. To simplify further discussion, we refer to any applicable similarity measure or distance metric as simply a similarity measure.

Popular post hoc XAI methods like \textsc{Lime}~\parencite{LIME_Ribeiro} 
provide a ranked list of features ordered by their importance with respect to the desired model output, and comparisons between ranked lists is a standard task in the field of information retrieval. In particular, previous work has extended commonly used measures for ranked lists to allow weighted similarity of elements \parencite{GenDistKumar,Sculley2007RankAF}. However, it is important to note that the use of these specific measures is not required to apply synonymity-weighted similarity. 
\subsection{Adversarial Attack Process}
Since the genesis of this work concerns adversarial attacks on XAI methods, we briefly reiterate the general adversarial attack process.

Let $\mathbf{M}$ be a model that can be explained using the XAI method $\mathbf{E}$. Let $\mathbf{I}$ be some input to $\mathbf{M}$ we desire the explanation to be centered upon, that is we desire to understand how $\mathbf{M}$ uses the components of $\mathbf{I}$ to produce some output. $\mathbf{M(I)}$ is fed into $\mathbf{E}$ to generate an initial explanation $\mathbf{A}$ used as the basis of comparison against future perturbed versions of $\mathbf{I}$, termed $\mathbf{B}$. 

The input $\mathbf{I}$ is then perturbed by replacing some subset, usually a single word, with an appropriate substitution. This substitution is a close synonym, as judged with respect to some measure, and is subject to certain constraints such as the grammatical category or location in the unperturbed explanation $\mathbf{I}$. The goal here is to generate a replacement while maintaining the fundamental meaning and structure of $\mathbf{I}$. 
The perturbed explanation $\mathbf{B}$ is compared to $\mathbf{I}$ using some similarity measure, and the process repeats with $\mathbf{B}$ as the new document to be perturbed until some termination condition. 

We note that prior work in adversarial XAI has enforced a single word to single word replacement for the perturbations due to the extensive reduction in search complexity and thus computation time (which remains a significant bottleneck) 
This restriction to a single word replacement scheme imposes an implicit mapping which can then be used to implement synonymity weighted similarity. 




\section{Mappings Between Explanations}\label{sec:mappings}
As our goal is to include the synonymity estimate between paired features $a \in \mathbf{A},b \in \mathbf{B}$ where $\mathbf{A},\mathbf{B}$  are the original and perturbed explanations, respectively, we require a way to determine the pairing $a \to b$ to allow comparison. To do so, we will define a mapping using the perturbation process to link elements from both lists.

Creating this mapping is simple if $|\mathbf{A}| = |\mathbf{B}|$ and the perturbation process is restricted to single-word substitutions. Any element $a \in \mathbf{A}$ located at index $i$ in $\mathbf{A}$ that is perturbed to some value $b$ must either be located at some index $j$ in $\mathbf{B}$ or no longer part of the surrogate model and so is missing from explanation $\mathbf{B}$. For the latter case, this generally means eliminated in importance. 
However, if the search process does not exclude highly ranked features, these can be selected (and often are without careful choice of search constraints), resulting in substantial differences under certain similarity measures despite little substantial difference in meaning. This is one of the major issues that we seek to address with synonymity weighting. 

We note that most perturbed features from standard text XAI perturbation methods remain present in the perturbed explanation, but when constraining the size of the explanation either through the surrogate model itself, or in the case of truncating the resulting explanation to the top-k features (done usually to reduce explanation complexity for the end-user) we often encounter unpaired elements between the explanations. For measures that rely on consonant lists, an adjustment must be made to allow for dissonant elements. In this case to handle any unpaired elements we apply a penalty value $p$, its value dependent on both the similarity measure and user choice. 

Some measures can be simply extended to handle disjoint elements. For example, Kendall's Tau (Rank Distance) which counts the number of dissonant pairs between two ranked lists, can be easily extended to the comparison of different sized lists simply by declaring whatever excess element(s) that exist in the larger list to be automatically dissonant. But even for measures whose structure discards element pairings (generally set intersection-based), we must establish a mapping to apply the synonymity weighting. 

We will assume that all elements contained within the original explanation $\mathbf{A}$ are mapped, either to an element in $\mathbf{B}$ or the null mapping which indicates that the measure specific penalty should be applied. Additionally, multi-word substitutions for the perturbation method increases the difficulty of maintaining appropriate syntactic consistency and have not seen much use in adversarial explanations in XAI. As such, we will assume that all perturbation methods replace at most one word per iteration. 

Now given our mapping, we must decide what form of synonymity weighting to apply.

\section{Constructing Weighted Similarity}\label{sec:weightingMethods}
With some mapping established between $\mathbf{A}$ and $\mathbf{B}$, how should the synonymity weighting be implemented? We consider two intuitive possibilities, one where the synonymity estimate is within an interval and the second where synonymity is a dichotomy.

For the interval estimate, we define a function $Syn(a,b) \longrightarrow [0,1]$ where $a, b$ are features within explanations $\mathbf{A}, \mathbf{B}$ respectively. The function $Syn(\cdot)$ returns a value proportional to the synonymity between the features $a$ and $b$. We constrain the definition of $Syn(a,b)$ minimally, with the only condition required being $Syn(x,x) = 1$, where $1$ is the absolute maximum similarity possible. In particular, the interval itself is subject to alteration. We choose $[0,1]$ to provide a simple representation for the proportion of similarity between two words and so make the analysis easier to perform, but this is not required and other choices are possible. Of these alternative choices, the interval $[-1,1]$ may be the most intuitive choice by letting $-1$ indicate that $b$ is an antonym of $a$, and $0$ being that $b$ is completely unrelated to $a$. 

For example, let $a = good$, $b = bad$, and $c = frog$. Now for the interval $[0,1]$, $Syn(a,b) = Syn(a,c) = 0$ as neither word is synonymous with \textit{good}. For the interval $[-1,1]$ $Syn(a,b) \approx -1$ and $Syn(a,c) = 0$ as \textit{bad} is an antonym of the adjective \textit{good}. Since part of speech checking is a commonly imposed constraint in text-based adversarial XAI, we assume that there is no ambiguity between words that span multiple syntactic categories such as \textit{ good} (\textbf{noun}) and \textit{good} (\textbf{adjective}) should the synonymity measure be capable of handling this distinction. In particular, measures that incorporate embedding vectors do not necessarily make a distinction between identical words with multiple meanings (the embeddings are not \textit{ multisense}). These embeddings also will often not produce values with such obvious delineation. \textit{Good} and \textit{frog} will likely possess some similarity $> 0$ by construction of the embedding. Choosing an optimal embedding is likely task specific and outside the scope of this discussion.

We can also represent synonymity with a method akin to that of a traditional thesaurus, that is, there is a fixed collection of synonyms for a given word with any other word by definition not being synonymous. For a given word $a$, define the set of valid synonyms for $a$ as $\mathbf{S}_a$. Then our similarity function $Syn(a,b)$ is the characteristic function $\chi({\mathbf{S}_a}(b))$ where if $b \in \mathbf{S}_a$ then $Syn(a,b) = 1$, otherwise it is $0$. This  ``thesaurus'' method may prove appealing to those who are dubious of the quality of estimates produced by comparison between word embeddings. This strategy is particularly useful for adversarial XAI methods whose search process focuses on perturbing important features first, as the change in similarity between adversarial substitutions can be unreasonably high for methods when large weights are applied to the top subset of features.

While both possibilities for synonymity weighting are reasonable, each comes with some amount of subjectivity. For the interval representation, the continuous-valued output retains some measure of comparison between synonyms, as not all synonyms may be truly identical substitutions. However, the question of how closely related are the words \textit{good} and \textit{great} is subject to personal interpretation. Of course, words will generally be represented in terms of some embedding, but different embeddings may have different representations of a word and subsequently different calculations for the resulting similarity. Defining a set of valid synonyms avoids the subjectivity associated with the numerical representation of the synonym, but the question now shifts as to why certain elements are contained within the set. 

For our main experiments, we selected the widely-used GloVe-Twitter-25 model to generate similarity scores. The results of this primary analysis are detailed in the subsequent sections (and seen Tables 3-4, Figures 1-2). Then, to test the robustness of our conclusions to choice of embedding we conduct a sensitivity analysis, replicating our experiments with two methodologically different options: fastText, which leverages subword information, and a thesaurus-based method derived from WordNet (as described above). As our results will show, this analysis demonstrates that the core findings of our work are consistent across all three measures, reinforcing the general applicability of our proposed weighting method.

\subsection{An Example: The Jaccard Index}
Here we demonstrate the idea by constructing weighted similarity using one of the simplest measures for the comparison of lists, the Jaccard Index. The Jaccard index is simply the ratio of the size of the nonempty intersection of the lists (here viewed as sets) to the size of their union (Eq. \ref{Jaccard}). If the intersection is empty, the resulting similarity is defined to be zero. XAI explanations tend to report only unique features, and so we assume that there are no duplicates within the explanations. However, this is not a strict requirement as the Jaccard index can be extended to multisets.

\begin{equation}\label{Jaccard}
    J(\mathbf{A},\mathbf{B}) = \frac{|\mathbf{A} \cap \mathbf{B}|}{|\mathbf{A} \cup \mathbf{B}|}
\end{equation}

\noindent Here, $\mathbf{A}$ is the original explanation and $\mathbf{B}$ is the perturbed explanation. 
\newline

Now consider the explanations $\mathbf{A} = \{a, b, c\} $ and $\mathbf{B} = \{\alpha, \beta, \gamma\} $. We assume that there exists some one-to-one mapping, $M(a_i) \to b_j$ with $i,j$ the indices of the elements $a,b$ in the explanations $\mathbf{A},\mathbf{B}$, respectively, imposed via the perturbation process from the elements of $\mathbf{A}$ to the elements of $\mathbf{B}$ as in Section \ref{sec:mappings}. Let this mapping be defined by $M(a_1) \to (\alpha_1)$, $M(b_2) \to (\beta_2)$, and $ M(c_3) \to (\gamma_3)$. Then the Jaccard Index gives us:
\begin{equation*}
\begin{gathered}
    J(\mathbf{A},\mathbf{B}) = \\
    \frac{| \{a, b, c\} \cap \{\alpha, \beta, \gamma \}|}{|\{\{a, 
    b, c\}\cup \{\alpha, \beta, \gamma \} \}|}  = \frac{\varnothing}{|\{a, b, c, \alpha, \beta, \gamma\ \}|} = 0
    \end{gathered}
\end{equation*} 
\newline

We can see that there is some level of correspondence between the previous two sets, despite the symbols being different. Both are the first three letters of their respective alphabets, while $a$ and $\alpha$ and $b$ and $\beta$ possess added similarity due to their functional equivalence as letters. The Jaccard index cannot capture any similarity by default. To remedy this, we define a similarity measure $\textit{Syn}(x,y) \to [0,1]$ that operates on elements $x \in \mathbf{A}$ and $y \in \mathbf{B}$. For equally sized lists of elements, we can rewrite Equation \ref{Jaccard} as:


\begin{align}\label{Jaccard2}
    J(\mathbf{A},\mathbf{B}) =  \frac{\displaystyle\sum_{i = 1}^{|\mathbf{A}|}|\mathbf{A}[i] \cap M(\mathbf{A}[i])|}{|\mathbf{A} \cup \mathbf{B}|} 
\end{align}

\noindent Then, applying our similarity measure $\textit{Syn($\cdot$)}$ we have:
\begin{align}\label{Jaccard3}
    J_W(\mathbf{A},\mathbf{B}) = \frac{\displaystyle\sum_{i = 1}^{|\mathbf{A}|}\textit{Syn}(\mathbf{A}[i],M(\mathbf{A}[i]))}{|\mathbf{A} \cup \mathbf{B}|} 
\end{align}



\noindent Now, define $\textit{Syn}(a,\alpha){=}0.9$, $\ \textit{Syn}(b,\beta){=}0.6$, and $ \ \textit{Syn}(c,\gamma){=}0.3$. Then 
\begin{equation*}
    J_W(\mathbf{A},\mathbf{B}) = \frac{0.9 + 0.6 + 0.3}
{|\mathbf{A} \cup \mathbf{B}|} = 0.3
\end{equation*}
\newline

Given that we have established a mapping between the elements, treating them as disjoint may be inappropriate as there exists some relation between the paired elements. We would then expect the similarity between them to be greater. To adjust for this, we can alter the denominator by the number of each pair of mapped elements. Then we have:
\begin{equation*}
  J_W(\mathbf{A},\mathbf{B}) = \frac{0.9 \ + \  0.6 \ + \  0.3}{|\mathbf{A} \cup \mathbf{B}| \ - \ 3} = 0.6  
\end{equation*}




In this example, we had both a mapping between every element and a total replacement of every element in the original explanation. In practice, this is unlikely. A complete replacement of each feature within an explanation requires extensive perturbations, usually requiring most of the document to be perturbed. And as the perturbation process is not flawless, this inevitably results in the severe degradation of the textual quality of the document, defeating the purpose of the stability testing.

\section{Empirical Validation} 
\label{sec:experiment}
To demonstrate the effectiveness of synonymity weighting, we modify four common similarity measures used for the comparison of collections of features in adversarial XAI. 
The formulations are defined below in Section \ref{measure_def}. 

\subsection{Similarity Measures}\label{measure_def} 
For the following definitions, let $\mathbf{A}, \mathbf{B}$ denote the ranked lists. If $|\mathbf{A}| \neq |\mathbf{B}|$, then without loss of generality assume $|\mathbf{A}| > |\mathbf{B}|$.
\newline

\noindent \textit{(1)  Jaccard Index} which has already been defined and discussed in Section \ref{sec:motivation}.
\newline

\noindent \textit{(2)  Kendall's Tau Rank Distance} (Eq. \ref{eq:Kendall}) counts the number of pairwise inversions between $\mathbf{A}$ and $\mathbf{B}$ (where $\bold{1}[\cdot]$ is the indicator function. To allow for comparison of lists unequal in size, we assume that all excess elements of the larger list are automatically disjoint.
\begin{equation}\label{eq:Kendall}
    \sum_{i=1}^{\textbf{min}(|\mathbf{A}|,|\mathbf{B}|)} (\bold{1}[\mathbf{A}[i] \neq \mathbf{B}[i]]) +  | |\mathbf{A}| - |\mathbf{B}| |
\end{equation}


We extend Kendall's Tau to use synonymity weighting by adjusting the value of a mapped dissonant pair $a,b$ at an equal location by multiplying by $1 - Syn(a,b)$. For highly synonymous replacements ($Syn(a,b) \to 1$) this assigns a distance close to zero for the mapped pair, and for dissimilar words it approaches the default distance. 

\begin{equation}\label{eq:Kendall2}
\begin{matrix} 
    \sum_{i=1}^{\textbf{min}(|\mathbf{A}|,|\mathbf{B}|)} (\bold{1}[\mathbf{A}[i] \neq \mathbf{B}[i]]*(1 - Syn(a,b))  \\
    \\
    + | |\mathbf{A}| - |\mathbf{B}| | 
    \end{matrix}
\end{equation}
\newline

\noindent \textit{(3)   Spearman's footrule} (Eq. \ref{eq:Spearman}) is the sum of the difference between the location $i$ of each feature $a \in \mathbf{A}$ to its corresponding location $j$ in $\mathbf{B}$. Spearman's footrule is effectively the $L_1$ distance between ranked lists. Like Kendall's Tau, the footrule is by default not intended for disjoint lists but can be altered by applying a penalty $p$ for disjoint elements. Here we use the formulation with a location parameter from \cite{topklists} which is designed for top k lists and choose a penalty value of $\frac{k}{2}$. 

\begin{equation}\label{eq:Spearman}
  \sum_{a \in \mathbf{A}} | i - j |
\end{equation}

We extend the footrule to use synonymity weighting by taking the summation of three mutually exclusive conditions that a pair of features within the explanations may have. For features unchanged in both explanations, we calculate the distance as normal. For features $a,b$ with a mapping $a \to b$ we calculate the minimum between the standard distance divided by the synonymity between features $Syn(a,b)$ and the maximum possible distance under the default footrule $|A| - 1$. 

\begin{equation}\label{eq:Spearman2}
  \sum_{a \in \mathbf{A\cap B}} | i - j | +  \sum_{\substack{a \in \mathbf{A} \\ b \in \mathbf{B} \\ a \to b}} s + \sum_{a \in \mathbf{A \cap \bar{B}}} p 
\end{equation}

\noindent Where $s = \mathbf{min}(\frac{| i - j | }{Syn(a,b)}, |\mathbf{A}| - 1)$
\newline

\noindent \textit{(4)  Rank-biased Overlap (RBO)} is a sum of successively larger intersections, each weighted by a term in a convergent series. This weighting scheme is controlled by a parameter $p \in (0,1)$ that can be adjusted to ascribe more or less weight to the top k features. In general, values further down the list are weighted as less significant, which is often the case in XAI, as only the top few features are of interest to many end users~\parencite{verma2020counterfactual}.
RBO \parencite{10.1145/1852102.1852106} is defined in Eq. \ref{eq:RBO} where 
 $d$ is the current depth of the ranking, and $k$ is the maximum depth. 
\begin{equation}\label{eq:RBO}
  RBO(\mathbf{A},\mathbf{B},p) =  (1-p)  \sum_{d = 1}^{k}{p^{d-1}}  \frac{|\mathbf{A}_{:d} \cap \mathbf{B}_{:d}|}{d}
\end{equation}
To enable weighted synonymity we apply the same idea as in Equations \ref{Jaccard2} and \ref{Jaccard3} where the size of the intersection is increased by the similarity of each mapped pair of disjoint elements.

\begin{table}[t]
    \caption{Comparative similarities on the explanation of an adversarial example in the S2D dataset}

    \footnotesize

    \begin{center}
     \textul{\textbf{Original Text} }  
    \end{center}
    
    I have a lot of heartburn and I feel like I'm choking when I eat. I 
    
    also have a lot of stomach pain and I vomit a lot.
    \newline
    
    \begin{center}
     \textul{\textbf{Perturbed Text}}   
    \end{center}

    I \textbf{\textit{got}} a \textbf{\textit{batch}} of \textbf{\textit{indigestion}} and I feel like I'm choking when I eat. 
    
    I also have a lot of stomach pain and I vomit a lot.
\vspace{10pt}

    \begin{subtable}[t]{.48\textwidth}
    \raggedright
    \setlength\tabcolsep{11 pt}
        \begin{tabular}{clc|clc}
           \multicolumn{3}{c}{\textbf{\textul{Original Explanation}}} & \multicolumn{3}{c}{\textbf{\textul{Perturbed Explanation}}} 
            \\
             \multicolumn{2}{c}{\textbf{}} & \multicolumn{1}{c}{\textbf{}}&  \multicolumn{2}{c}{\textbf{}} & \multicolumn{1}{c}{\textbf{}}  \\
            \hline \\
            1 & heartburn &  & 1 & choking &   \\
                2 & eat &  & 2 & vomit &  \\
                3 & vomit &  & 3 & eat &  \\
                4 & choking &  & 4 & batch &  \\
                5 & stomach &  & 5 & indigestion &  \\
                6 & feel &  & 6 & pain &  \\
                 & ... & &  & ...  & \\
                
            \hline \\
        \end{tabular}
        \end{subtable}

\renewcommand{\tabcolsep}{2.5pt}
\centering
\begin{subtable}[t]{.48\textwidth}
\caption*{\textbf{Comparative Similarities}}
\raggedright
\begin{tabular}{cccccc}
\toprule

\multicolumn{6}{c}{\textbf{\textul{Standard}}} \vspace{4pt} \\

\textbf{Jaccard} & \textbf{Kendall} & \textbf{Spearman} & \textbf{RBO$_{0.5}$} & \textbf{RBO$_{0.7}$} & \textbf{RBO$_{0.9}$}\\
0.50 & 0.11 & 0.39 & 0.16 & 0.31 & 0.51

\\

\multicolumn{6}{c}{\textbf{\textul{Weighted}}} \vspace{4pt} \\
\textbf{Jaccard} & \textbf{Kendall} & \textbf{Spearman}
& \textbf{RBO$_{0.5}$} & \textbf{RBO$_{0.7}$} & \textbf{RBO$_{0.9}$}\\
0.62 & 0.11 & 0.61 & 0.17 & 0.32 & 0.56\\

\bottomrule
\end{tabular}
    \end{subtable}%
    \label{tab:comparative_perturbations_example_2}
\end{table}

\newcolumntype{Y}{>{\centering\arraybackslash}X}
\begin{table*}[ht]
\caption{Successful attack rates under threshold $\tau$ for standard and synonymity weighted explanations}
\footnotesize
\begin{tabularx}{\textwidth}{c *{13}{Y}}
\toprule
\multicolumn{1}{c}{\textbf{}}
& \multicolumn{1}{c}{\textbf{$\tau$}}
& \multicolumn{2}{c}{\textbf{Jaccard}}
& \multicolumn{2}{c}{\textbf{Kendall}}
& \multicolumn{2}{c}{\textbf{Spearman}}
& \multicolumn{2}{c}{\textbf{RBO}$_{0.5}$}
& \multicolumn{2}{c}{\textbf{RBO}$_{0.7}$}
& \multicolumn{2}{c}{\textbf{RBO}$_{0.9}$}\\
\cmidrule(lr){2-14}
 & & \textbf{Base} & \textbf{Syn$_w$} & \textbf{Base} & \textbf{Syn$_w$} & \textbf{Base} & \textbf{Syn$_w$} & \textbf{Base} & \textbf{Syn$_w$} & \textbf{Base} & \textbf{Syn$_w$} & \textbf{Base} & \textbf{Syn$_w$}
 \\ \addlinespace[3pt]
 & 30\%  & 0.02 &  0 & 0.95 & 0.88 & 0.14 & 0 & 0.12  & 0.10   & 0 & 0 &  0 &  0 \\
 \rotatebox[origin=c]{90}{\textbf{GB}}& 40\%   &  0.24 & 0  & 0.98 & 0.90 & 0.38  & 0 & 0.26 & 0.17  & 0.07 &  0.05 & 0 & 0 \\
 & 50\%  & 0.88 &  0 &  1 & 0.86 & 0.83 & 0.02 & 0.40 & 0.40   & 0.28 & 0.14 &  0.07 &  0  \\
 & 60\%  & 1 &  0.05 &  1 & 0.88 & 1 & 0.05 & 0.40 & 0.40    & 0.40 & 0.33 &  0.28 &  0 \\
   \midrule
 & 30\%  & 0.06 & 0  &  1 & 0.94 & 0.18 & 0 & 0.06 & 0.06   & 0.02 & 0 & 0 &  0 \\
 \rotatebox[origin=c]{90}{\textbf{S2D}}& 40\%  & 0.52 &  0.06 &  1 & 0.94 & 0.58 & 0  & 0.18 & 0.14   & 0.04 & 0.04 &  0 &  0 \\
 & 50\%  & 0.98 &  0.08 &  1 & 0.96 & 0.92 & 0 & 0.24 & 0.24   & 0.20 & 0.12 &  0.08 &  0 \\
 & 60\%  & 1 &  0.18 &  1 & 1 & 1 & 0 & 0.24 & 0.24   & 0.30 & 0.26 & 0.28 & 0.10\\
\bottomrule
\end{tabularx}

\label{tab:success}
\end{table*}

\begin{figure*}[!htb]

\minipage{\textwidth}
  \centerline{\includegraphics[width=\linewidth,height=0.225\linewidth]{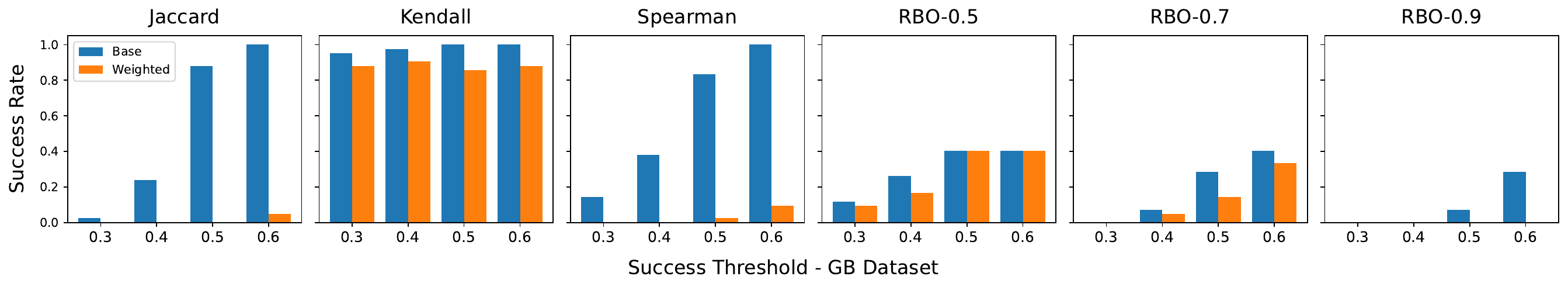}}
  \endminipage
\hfill
\minipage{\textwidth}  \centering
  \centerline{\includegraphics[width=\linewidth,height=0.225\linewidth]{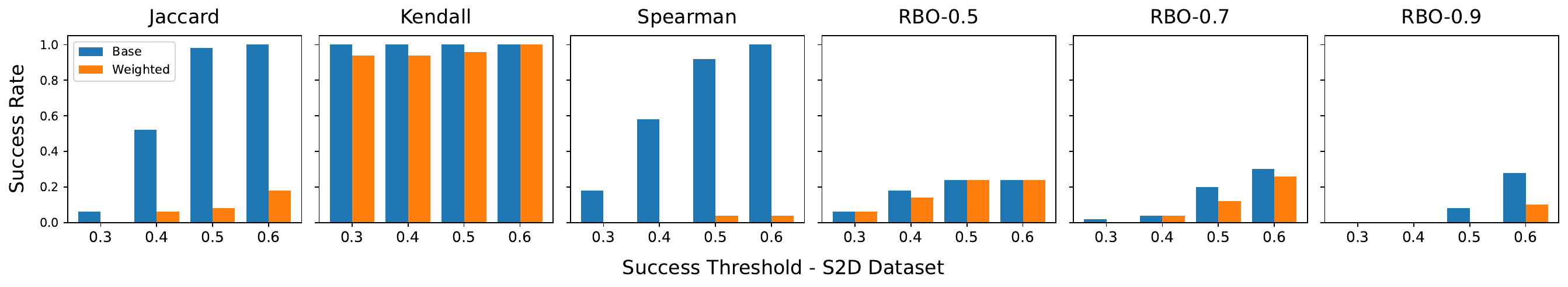}}
  \endminipage
\caption{Successful attack rates under threshold $\tau$ for standard and synonymity weighted explanations} 
\label{fig:success}%
\vspace{-10pt}
\end{figure*}

\subsection{Experimental Data}
To test the effects of synonymity weighting on stability estimates, we require examples generated from an adversarial XAI process. We use the method in \cite{burger2023explanations} that easily allows one to replace the similarity measure used to guide the search process. As the similarity measure used for the comparison of explanations controls the search, simply comparing the end result calculated under a single measure will not accurately describe the stability, as the replacement of words with viable perturbations can differ appreciably between measures used. 

We note that the attack process itself is unmodified in that synonymity weighting is not applied to the measure guiding the attack but only to the final output. This was done to prevent unnecessary computation as repeating the attack with the added synonymity weighting would have required over a month of additional computation on an A6000 GPU. This renders our conclusions \textit{understated}, as our construction of the synonymity weighting results, at worst, in the same similarity to the original measure. For all practical purposes, the similarity will increase, which will reduce the attack success rate.

To create the raw material for our analysis, we generate batches of 50 adversarial examples using the algorithm in \cite{burger2023explanations}. Each batch is generated with respect to a similarity measure and a success threshold. The four measures defined in Section \ref{measure_def} are used: The Jaccard index, Kendall's Tau Rank Distance, and Spearman's footrule, and RBO with weighting parameters 0.5, 0.7, and 0.9. Combined with the thresholds of 30\%, 40\%, 50\%, and 60\%, this results in 1,200 adversarial examples per data set. For the data sets, we use two of those included in \cite{burger2023explanations} and their associated pre-trained DistilBERT models. The first data set is the Twitter dataset with short length (average 11 words) gender bias Twitter dataset (\textbf{GB}) and the second is the data set of symptoms to diagnosis of moderate length (average 29 words) symptoms to diagnosis dataset (\textbf{S2D}).

\begin{table*}[h]
\caption{Successful attack similarity levels before and after synonymity weighting} 
\footnotesize
\begin{tabularx}{\textwidth}{c *{13}{Y}}
\toprule
\multicolumn{1}{c}{\textbf{}}
& \multicolumn{1}{c}{\textbf{$\tau$}}
& \multicolumn{2}{c}{\textbf{Jaccard}}
& \multicolumn{2}{c}{\textbf{Kendall}}
& \multicolumn{2}{c}{\textbf{Spearman}}
& \multicolumn{2}{c}{\textbf{RBO}$_{0.5}$}
& \multicolumn{2}{c}{\textbf{RBO}$_{0.7}$}
& \multicolumn{2}{c}{\textbf{RBO}$_{0.9}$}\\
\cmidrule(lr){2-14}
& & \textbf{Base} & \textbf{Syn$_w$} & \textbf{Base} & \textbf{Syn$_w$} & \textbf{Base} & \textbf{Syn$_w$} & \textbf{Base} & \textbf{Syn$_w$} & \textbf{Base} & \textbf{Syn$_w$} & \textbf{Base} & \textbf{Syn$_w$}
\\ \addlinespace[3pt]
& 30\%  & 0.27 &  0.51 & 0.19 & 0.21 & 0.24 & 0.51 & 0.28  & 0.28  & - & - &  - &  - \\
\rotatebox[origin=c]{90}{\textbf{GB}}& 40\%   &  0.39 & 0.56  & 0.26 & 0.27 & 0.33  & 0.58 & 0.34 & 0.34   & 0.38 & 0.40 & - & - \\
& 50\%  & 0.46 &  0.61 &  0.32 & 0.33 & 0.45 & 0.68 & 0.40 & 0.40  & 0.47 & 0.50 &  0.46 &  0.58  \\
& 60\%  & 0.55 &  0.68 &  0.37 & 0.39 & 0.53 & 0.72 & 0.40 & 0.40    & 0.52 & 0.54&  0.57 &  0.67 \\
\midrule
& 30\%  & 0.29 & 0.42  &  0.22 & 0.23 & 0.26 & 0.59 & 0.27 & 0.27   & 0.28 & 0.31 & - &  - \\
\rotatebox[origin=c]{90}{\textbf{S2D}}& 40\%  & 0.37 &  0.49 &  0.29 & 0.30 & 0.36 & 0.65  & 0.35 & 0.35   & 0.40 & 0.40 &  - &  - \\
& 50\%  & 0.47 &  0.57 &  0.36 & 0.37 & 0.46 & 0.70 & 0.42 & 0.42   & 0.48 & 0.49 &  0.49 &  0.57 \\
& 60\%  & 0.56 &  0.65 &  0.38 & 0.39 & 0.55 & 0.74 & 0.42 & 0.42 & 0.55   & 0.56 & 0.58 & 0.62 \\
\bottomrule
\end{tabularx}

\label{tab:sim}
\end{table*}

\begin{figure*}[!htb]

\minipage{\textwidth}
  \centerline{\includegraphics[width=\linewidth,height=0.225\linewidth]{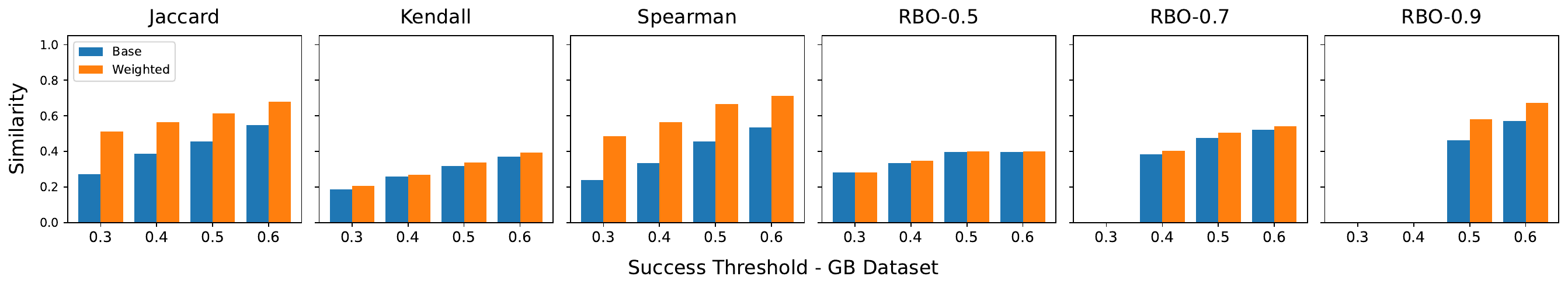}}
  \endminipage
\hfill
\minipage{\textwidth}  \centering
  \centerline{\includegraphics[width=\linewidth,height=0.225\linewidth]{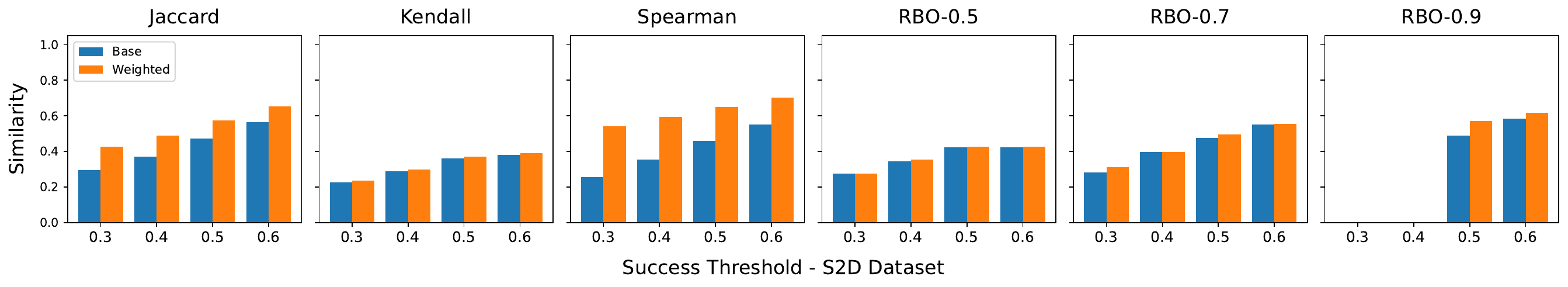}}
  \endminipage
  \caption{Successful attack similarity levels before and after synonymity weighting} 
\label{fig:sim}%
\vspace{-10pt}
\end{figure*}

\begin{table*}[h] 
\caption{Comparative average success rates across methods for measuring similarity}
\label{tab:sensitivity_average_success}
\footnotesize
\begin{tabularx}{\textwidth}{l l *{12}{Y}}
\toprule
 &  & \multicolumn{2}{c}{\textbf{Jaccard}} & \multicolumn{2}{c}{\textbf{Kendall}} & \multicolumn{2}{c}{\textbf{Spearman}} & \multicolumn{2}{c}{\textbf{RBO$_{0.5}$}} & \multicolumn{2}{c}{\textbf{RBO$_{0.7}$}} & \multicolumn{2}{c}{\textbf{RBO$_{0.9}$}} \\
\cmidrule(lr){3-4} \cmidrule(lr){5-6} \cmidrule(lr){7-8} \cmidrule(lr){9-10} \cmidrule(lr){11-12} \cmidrule(lr){13-14}
 &  & \textbf{Base} & \textbf{Syn$_w$} & \textbf{Base} & \textbf{Syn$_w$} & \textbf{Base} & \textbf{Syn$_w$} & \textbf{Base} & \textbf{Syn$_w$} & \textbf{Base} & \textbf{Syn$_w$} & \textbf{Base} & \textbf{Syn$_w$} \\
\midrule
\multirow{3}{*}{\rotatebox[origin=c]{90}{\textbf{GB}}} & \textbf{GloVe (Original)} & 0.54 & 0.01 & 0.98 & 0.88 & 0.59 & 0.03 & 0.30 & 0.27 & 0.19 & 0.13 & 0.09 & 0.00 \\
\addlinespace[2pt]
 & \textbf{fastText} & 0.54 & 0.02 & 0.98 & 0.88 & 0.59 & 0.04 & 0.30 & 0.27 & 0.19 & 0.13 & 0.09 & 0.01 \\
\addlinespace[2pt]
 & \textbf{WordNet} & 0.54 & 0.34 & 0.98 & 0.93 & 0.59 & 0.00 & 0.30 & 0.29 & 0.19 & 0.18 & 0.09 & 0.05 \\
\midrule
\multirow{3}{*}{\rotatebox[origin=c]{90}{\textbf{S2D}}} & \textbf{GloVe (Original)} & 0.64 & 0.08 & 1.00 & 0.96 & 0.67 & 0.02 & 0.18 & 0.17 & 0.14 & 0.11 & 0.09 & 0.03 \\
\addlinespace[2pt]
 & \textbf{fastText} & 0.64 & 0.07 & 1.00 & 0.95 & 0.67 & 0.02 & 0.18 & 0.17 & 0.14 & 0.12 & 0.09 & 0.03 \\
\addlinespace[2pt]
 & \textbf{WordNet} & 0.64 & 0.28 & 1.00 & 0.98 & 0.67 & 0.01 & 0.18 & 0.17 & 0.14 & 0.12 & 0.09 & 0.08 \\
\bottomrule
\end{tabularx}
\end{table*}

\begin{table*}[h] 
\caption{Comparative average similarity levels across methods for measuring similarity}
\label{tab:sensitivity_average_similarity}
\footnotesize
\begin{tabularx}{\textwidth}{l l *{12}{Y}}
\toprule
 &  & \multicolumn{2}{c}{\textbf{Jaccard}} & \multicolumn{2}{c}{\textbf{Kendall}} & \multicolumn{2}{c}{\textbf{Spearman}} & \multicolumn{2}{c}{\textbf{RBO$_{0.5}$}} & \multicolumn{2}{c}{\textbf{RBO$_{0.7}$}} & \multicolumn{2}{c}{\textbf{RBO$_{0.9}$}} \\
\cmidrule(lr){3-4} \cmidrule(lr){5-6} \cmidrule(lr){7-8} \cmidrule(lr){9-10} \cmidrule(lr){11-12} \cmidrule(lr){13-14}
 &  & \textbf{Base} & \textbf{Syn$_w$} & \textbf{Base} & \textbf{Syn$_w$} & \textbf{Base} & \textbf{Syn$_w$} & \textbf{Base} & \textbf{Syn$_w$} & \textbf{Base} & \textbf{Syn$_w$} & \textbf{Base} & \textbf{Syn$_w$} \\
\midrule
\multirow{3}{*}{\rotatebox[origin=c]{90}{\textbf{GB}}} & \textbf{GloVe (Original)} & 0.42 & 0.59 & 0.28 & 0.30 & 0.39 & 0.61 & 0.35 & 0.36 & 0.34 & 0.36 & 0.26 & 0.31 \\
\addlinespace[2pt]
 & \textbf{fastText} & 0.42 & 0.57 & 0.28 & 0.30 & 0.39 & 0.61 & 0.35 & 0.36 & 0.34 & 0.36 & 0.26 & 0.30 \\
\addlinespace[2pt]
 & \textbf{WordNet} & 0.42 & 0.51 & 0.28 & 0.29 & 0.39 & 0.79 & 0.35 & 0.35 & 0.34 & 0.35 & 0.26 & 0.28 \\
\midrule
\multirow{3}{*}{\rotatebox[origin=c]{90}{\textbf{S2D}}} & \textbf{GloVe (Original)} & 0.43 & 0.54 & 0.31 & 0.32 & 0.41 & 0.62 & 0.37 & 0.37 & 0.43 & 0.44 & 0.27 & 0.30 \\
\addlinespace[2pt]
 & \textbf{fastText} & 0.43 & 0.53 & 0.31 & 0.32 & 0.41 & 0.63 & 0.37 & 0.37 & 0.43 & 0.43 & 0.27 & 0.29 \\
\addlinespace[2pt]
 & \textbf{WordNet} & 0.43 & 0.48 & 0.31 & 0.32 & 0.41 & 0.71 & 0.37 & 0.37 & 0.43 & 0.43 & 0.27 & 0.28 \\
\bottomrule
\end{tabularx}
\end{table*}

\section{Results \& Discussion}

All the following results are calculated exclusively with respect to successful attacks. We provide data for before and after the application of synonymity weighting on the overall success rate of the attack given some success threshold $\tau$ (Figure \ref{fig:success}, Table \ref{tab:success}) and the average resulting similarity of the perturbed explanation to the original at the end of the attack process. (Figure \ref{fig:sim}, Table \ref{tab:sim}). The method used to determine synonymity is cosine similarity on the pre-trained GloVe Twitter embedding. 
\newline

\noindent \textbf{Jaccard \& Spearman:} Immediately seen is the drastic reduction in the attack success rate for the Jaccard index and Spearman's footrule. For Jaccard under the \textbf{GB} dataset, 
we see success rates for the GB dataset under Jaccard fall to nearly zero for all but the most lenient (60\%) threshold.
Similar results hold for the \textbf{S2D} dataset. Spearman is even more sensitive than Jaccard, with every \textbf{S2D} success under the original calculation changed to failure. Spearman's footrule was investigated to determine if the weighting scheme was appropriate due to the enormous success reduction and comparatively more complex weighting formulation over Jaccard. However, the different choices of the penalty value showed little change in the results with only slight ($\leq 5\%$) reductions in calculated similarity. These results demonstrate that both standard measures are extremely sensitive to changes in the explanation that are due to properties of the measure, and not due to a change in fundamental meaning of the explanation. With such significant decreases in attack success, any conclusions generated about XAI stability under these measures should be viewed with suspicion. 
\newline


\noindent \textbf{RBO:} In contrast to Jaccard and Spearman, RBO gains less from synonymity weighting, only showing minor changes in success rate for weighting values of 0.5 and 0.7. For weighting value 0.9 it is not surprising to see more efficacy as a more uniform distribution of importance to each feature. There still exist important examples where RBO can profit through the use of synonymity weighting (Table \ref{tab:comparative_perturbations_example}), but in general the changes are modest (Table \ref{tab:comparative_perturbations_example_2}). The lack of overall success associated with RBO$_{0.9}$ makes a judgement on its usefulness premature without more data from successful attacks. The appreciable difference in success rate change between RBO and Jaccard was not expected as both are intersection-based. RBO's relatively minor gain from synonymity weighting is likely due to its own inherent weighting scheme; to confirm this, the intrinsically weighted version of the other measures can be tested, and we leave this for future investigation. Overall, RBO remains a strong candidate for use in adversarial XAI as its design innately maintains a balance between sensitivity and indifference.
\newline

\noindent \textbf{Kendall:} Kendall's Tau also appears to gain less from synonymity weighting, with minor changes on \textbf{S2D} and about twice the effectiveness on \textbf{GB}. This conclusion may be premature, as the extreme sensitivity (note the near 100\% success rate across every threshold and dataset) may overpower the effects of synonymity weighting. The construction of Kendall's Tau results in minor changes in ordering being shown as large deviations with the application of synonymity weighting in some cases noticeably increasing the resulting similarity (Table \ref{tab:comparative_perturbations_example}), but with final values well below most thresholds. The measure itself appears to be poorly suited for adversarial XAI and the synonymity weighting tested here is not sufficient to allow Kendall to see much, if any, practical use.



\subsection{Sensitivity Analysis} \label{sec:sensitvity}
To provide robustness in our conclusions about synonymity weighting we performed a sensitivity analysis that adds two methodologically different options: fastText, which leverages subword information, and a thesaurus-based method derived from WordNet. This analysis was conducted with respect to average success and similarity levels across the two datasets (Tables \ref{tab:sensitivity_average_success}-\ref{tab:sensitivity_average_similarity}).  

Noticeably, there is  little difference between the original GloVe embedding used and fastText, which is generally considered to be a superior alternative. The \textit{only} deviations between GloVe and fastText in the success rate analysis is 1\%, indicating that both embeddings are largely interchangeable under our method despite their different construction. The thesaurus-based option using WordNet showed more difference as expected due to its dichotomous construction where words are either synonyms and so are given full value (1) or not and given a value of (0). In general, the thesaurus-based method produced values closer to those calculated without synonymity weighting over the embedding based methods. However, most values were fairly close with only the Jaccard measure providing noteworthy mention. The success rates were substantially higher ($\sim32\%$ for GB and $\sim21\%$ S2D) compared to the embedding-based methods, but the overall success rate compared to the original base was far lower with a 20\% reduction on GB and a 36\% reduction on S2D. Overall, the choice of measure for generating the word-to-word similarity values does not appear to be particularly important and all choices tested result in superior estimates of XAI stability.

\section{Limitations and Conclusion}
The weighting constructions presented are intended as a proof-of-concept and may not be optimal. A limitation is that our analysis applies synonymity weighting post hoc; future work could integrate these weighted measures directly into the adversarial search process. This could be particularly beneficial for measures like Spearman's footrule, where a more integrated approach might mitigate the severe changes in similarity observed. Furthermore, while our choice of numerical synonymity is robust, superior estimates could be obtained using more advanced techniques, such as contextual embeddings from transformer models like BERT or by incorporating part-of-speech data. Finally, beyond improving the measures, future work could apply this robust evaluation framework to benchmark the stability of different XAI methods against each other, providing a fairer comparison of their stability.
  
Overall, this work demonstrates the usefulness of synonymity weighting to allow superior estimates of XAI stability. Substantial reductions in attack success rate for certain measures show the possibility of extremely understating XAI stability without the judicious selection of an appropriate adversarial perturbation method. The application of this weighting scheme also incurs negligible computational overhead as the bottleneck in current algorithms is the explanation generation itself, so even measures that see only moderate impact from synonymity weighting can incorporate it without burden. With the inclusion of synonymity in these measures, we offer a more robust method for threat detection in AI systems, ensuring that defensive efforts are focused on genuine vulnerabilities rather than measurement artifacts.
\\

\noindent \textbf{Data Availability Statement}: All data and materials used in this study are openly available in a public repository: {\small \url{https://github.com/christopherburger/SynEval}}
\\

\noindent \textbf{Acknowledgments}: The authors have no acknowledgments to declare.


\label{sec:end}





\printbibliography

\end{document}